\documentclass[runningheads]{llncs}
\usepackage[T1]{fontenc}
\usepackage{graphicx}
\usepackage{xcolor} 
\usepackage{booktabs}
\begin{document}

\title{Improving Model Safety \\ by Targeted Error Correction
\thanks{This work has been partially supported by MCIN/AEI/10.13039/501100011033 under the Mar\'ia de Maeztu Units of Excellence Program (CEX2021-001195-M).}
\thanks{This work has been accepted for publication in the Proceedings of International Conference on Pattern Recognition (ICPR) 2026. The final published version should be cited.}
}
%
%
\author{Abolfazl Mohammadi-Seif\inst{1}\orcidID{0000-0001-6616-433X}\\ \and
Ricardo Baeza-Yates\inst{1,2}\orcidID{0000-0003-3208-9778} %
}

\institute{
Universitat Pompeu Fabra, Barcelona, Spain
\and KTH Royal Institute of Technology, Stockholm, Sweden
\email{abolfazl.mohammadiseif@upf.edu, rbaeza@acm.org}}

%
\maketitle 
\begin{abstract}
The widespread adoption of machine learning in critical applications demands techniques to mitigate high-consequence errors. Our method utilizes a dual-classifier GBDT pipeline to distinguish routine human-like errors from high-risk non-human misclassifications. Evaluated across three domains, animal breed classification, skin lesion diagnosis (ISIC 2018), and prostate histopathology (SICAPv2), our framework demonstrates robust safety improvements. To address real-world deployment concerns, our results confirm the pipeline introduces negligible inference latency (1.60\% overhead for the animal dataset, 1.84\% for ISIC, and 1.70\% for SICAPv2) while outperforming traditional Maximum Class Probability (MCP) baselines in correction precision. Our conservative correction strategy successfully reduced dangerous non-human errors by 34.1\% in ISIC and 12.57\% in SICAPv2, improving super-class diagnostic safety to 90.41\% and 92.13\% respectively. This proves that safety-critical reliability can be substantially enhanced post-hoc without expensive model retraining.

\end{abstract}
%
%
\section{Introduction}

Machine learning (ML) models have achieved remarkable success in computer vision, often surpassing human performance in specific tasks. However, despite their high accuracy, these models remain prone to failures that differ fundamentally from human errors. When a human misclassifies an image, the mistake usually stems from visual ambiguity or fine-grained similarity (e.g., confusing a \textit{Siberian Husky} with a \textit{Malamute}), or structural data complexity~\cite{mohammadiseif2026cai}. In contrast, even deep neural networks can produce high-confidence predictions that are semantically incoherent to humans (e.g., confusing a \textit{dog} with a \textit{cat}, or a clearly \textit{benign} mole with a \textit{malignant} melanoma). These {\em non-human} errors ~\cite{baeza2022relevance} are not just accuracy penalties; they erode trust and, in safety-critical domains such as medical diagnosis or autonomous driving, pose severe risks ~\cite{omrani2022trust,raji2023concrete,ryan2020ai,mohammadiseif2026ijcnn}. In fact, in autonomous driving, most of the fatal cases are due to {\em non-human} errors such as not recognizing a white truck over a cloudy sky (Tesla) or a woman crossing at night in a bicycle (Uber).

The core issue lies in the fact that standard training goals, such as cross-entropy loss, treat all misclassifications equally. They penalize the model based on the probability assigned to the correct class, regardless of whether the predicted wrong class makes semantic sense. Consequently, a model might achieve high aggregate accuracy while harboring a propensity for catastrophic, {\em non-human} failures. In fact, evaluating success (e.g., accuracy) implies that all errors have the same impact, which is never true in practice.

To address this, we introduce a fully automated, post-hoc error correction technique that operates without access to ground truth labels during inference. Unlike prior oracle-guided approaches that assume knowledge of which predictions are incorrect, our method employs a realistic three-stage pipeline. First, a base multi-class classifier generates initial predictions.
Second, a {\em gatekeeper} model (Error Detector) analyzes the base model's output to predict the likelihood of an error (Correct vs. Error).
Third, if an error is detected, a {\em mitigation} model (Error Classifier) determines the nature of the error: is it a plausible {\em human-like} confusion or a potentially dangerous {\em non-human} failure?
Based on this diagnosis, we apply a targeted correction policy that flips {\em non-human} errors to a semantically safer superclass while retaining plausible {\em human-like} predictions.

We evaluate this approach on three distinct datasets: a custom in-the-wild animal dataset (trained on the Oxford-IIIT Pet dataset~\cite{parkhi2012cats}), the ISIC-2018 Skin Lesion Analysis dataset (critical benign vs. malignant diagnosis)~\cite{codella2019skin,tschandl2018ham10000}, and the SICAPv2 dataset (prostate histopathology)\cite{silva2020going}. The main contributions of this work are two-fold:
\begin{enumerate}
    \item \textbf{Automated Harm Mitigation Pipeline:} We propose a modular, model-agnostic cascade of classifiers in the context of multi-class classification. By inserting a learned Error Detector and Error Type Classifier after the base model, we can filter and correct predictions in real-time. This approach does not require costly retraining the large base backbone, making it computationally efficient. To address real-world deployment concerns, our measurements show an overhead of 0.1 ms per image, resulting in a latency increase of 1.60\%, 1.84\%, and 1.70\% for the animal, ISIC, and SICAPv2 datasets respectively. Furthermore, compared to a Maximum Class Probability (MCP) baseline, our error detector achieves higher precision (0.62, 0.66, and 0.49 respectively), preventing unwarranted corrections.
    \item \textbf{Empirical Validation in Three Use Cases:} We demonstrate that this pipeline improves safety as well as reliability. On the animal dataset, our method reduced {\em non-human} errors by approximately 7.7\% while improving overall class accuracy from 76.9\% to 77.6\%. On the ISIC-2018 medical dataset, the pipeline acted as a safety net, reducing dangerous diagnostic errors by 34.1\% while improving class accuracy from 79.2\% to 83.5\%. On the SICAPv2 dataset, the framework decreased non-human errors by 12.6\% and improved super-class safety accuracy from 91.00\% to 92.13\%.
\end{enumerate}

To guide our investigation into automated safety improvements, we formulate the following research questions:
\begin{itemize}
    \item \textbf{RQ1:} Can a secondary classifier effectively distinguish between correct predictions and misclassifications without access to ground truth labels during inference?
    \item \textbf{RQ2:} Once an error is detected, is it possible to automatically categorize it as a benign {\em human-like} confusion or a severe {\em non-human} failure based solely on the base model's output dynamics?
    \item \textbf{RQ3:} Does a targeted correction policy, triggered only by predicted {\em non-human} errors, improve the overall reliability and safety of the system, even when the error detection mechanism is imperfect?
\end{itemize}

The remainder of this paper is organized as follows. Section~\ref{sec:related} reviews related work. Section~\ref{sec:method} defines the error types and details our proposed error correction pipeline that addresses our three research questions, while Section~\ref{sec:results} presents the experimental results on the three use cases. Section~\ref{sec:discussion} discusses the real-world applicability, the error dynamics, and the limitations. Finally, Section~\ref{sec:Conclusion} concludes the work and outlines future research directions.

\section{Related Work}
\label{sec:related}

For a long time, accuracy was the gold standard for evaluating ML models. The logic was straightforward: a highly accurate model was considered successful. But over time, researchers began to highlight the limitations of relying solely on this metric. A model might achieve high general accuracy while still failing in meaningful and sometimes dangerous ways ~\cite{baeza2022relevance,foody2023challenges,lavazza2023common}. These failures often stem from fundamental differences between the way humans and machines process information. Human perception typically focuses on shapes and relationships between objects, whereas ML models often lock in more superficial features, such as textures~\cite{geirhos2020shortcut}. This divergence in perceptual strategy leads to systematic error patterns that a single accuracy score tends to obscure ~\cite{geirhos2018imagenet,patel2008investigating,ribeiro2016should}. As a result, there is growing recognition that we need evaluation frameworks and loss functions that go beyond accuracy or standard averages to account for how models succeed, and crucially, also how they fail~\cite{mohammadiseif2026bimodalloss}.

Equally important is how users perceive these failures. Studies show people respond much more negatively to unnatural mistakes, sparking a deeper distrust known as algorithm aversion~\cite{dietvorst2015algorithm}. These {\em non-human} errors, formalized by Baeza-Yates and Estévez-Almenzar~\cite{baeza2022relevance}, occur when models make decisions that defy common sense, such as confidently misclassifying a dog as a cat, which undermines their perceived intelligence and safety. Conversely, users are more forgiving when errors are predictable or resemble human reasoning~\cite{bansal2019beyond}. This reinforces that trust in AI does not hinge solely on strict correctness, but on whether its failures make sense within a shared human frame of reference.

Despite these insights, most current approaches stop at diagnosing the problem. Some strategies focus on flagging uncertain predictions for human review ~\cite{geifman2019selectivenet}, while others attempt to detect when models are operating outside their domain of competence, using mechanisms like trust scores ~\cite{jiang2018trust}. These are valuable tools, but they largely treat error as a static property, something to identify and avoid, rather than understand and address. Moreover, not all errors carry the same consequences, and traditional evaluations often overlook this variability, especially when it comes to {\em non-human} errors that reflect failures a human would rarely make. Our work advances the field by proposing a post-hoc correction technique that directly targets these high-risk, low-frequency mistakes. Rather than retraining the original model, we introduce a secondary model that learns to differentiate between routine, {\em human-like} misclassifications and more severe, {\em non-human} ones, such as confusing animals in image recognition, or malignant with benign in clinical diagnoses. This enables targeted intervention: we retain predictions for benign, familiar errors, while flipping or flagging those potentially harmful. In doing so, we offer a scalable, fast strategy to mitigate model risk in domains where certain errors could lead to irreversible outcomes. Ultimately, the approach doesn’t just reduce error, it reshapes it, aligning model behavior more closely with human expectations and real-world stakes.

To contextualize our approach, we now review related work on assessor models, which predict model behavior; representation learning methods, which address similar error challenges through improved feature learning; and the perception of users for different types of errors, highlighting how our post-hoc technique complements these strategies.

\paragraph{Assessor Models and Error Prediction.}
Our work aligns with the concept of assessor models~\cite{hernandez2022training}, which predict base model behavior on new instances. However, while assessor models typically focus on general performance prediction, our approach specifically targets the distinction between {\em human-like} and {\em non-human} error patterns, enabling targeted correction rather than just prediction.

\paragraph{Representation Learning Methods.}
Recent representation learning approaches like CLIP~\cite{radford2021learning} and DINO~\cite{caron2021emerging} learn discriminative features that naturally reduce inter-class confusions. These methods address similar issues through improved embeddings but require retraining or fine-tuning the base model. Our post-hoc technique complements these approaches by providing error correction without modifying the original classifier, making it suitable for deployed systems where retraining the base model is impractical.

\paragraph{Divergence of Human and Machine Errors.}
Prior research emphasizes that deep learning models often fail in ways that are fundamentally different from human cognition. For example, in the domain of face recognition, Estévez-Almenzar et al.~\cite{estevez2025comparison,estevezhuman2025} 
demonstrated that while human errors are typically driven by environmental factors like poor lighting or challenging angles, models are more prone to structural false positives, such as incorrectly identifying two different people as the same individual. These findings underscore that machine failures cannot always be interpreted through the lens of human perception, motivating the need for error correction techniques that explicitly distinguish between {\em human-like} and {\em non-human} errors.
\section{Methodology}
\label{sec:method}
Our approach shifts from a theoretical error analysis to a practical, fully automated pipeline capable of correcting model failures in real-time without access to ground truth labels. Our method consists in a pipeline that uses up to three models: (1) Base Classification, (2) Error Detection (distinguishing correct predictions from misclassifications), and (3) Error Type Classification (distinguishing plausible {\em human-like} errors from potentially dangerous {\em non-human} errors).

For the animal dataset, consistent with the definition of logical consistency failures in~\cite{baeza2022relevance}, we strictly defined {\em non-human} errors as cross-species misclassifications (e.g., cat predicted as dog). For ISIC and SICAPv2, {\em non-human} errors were defined as cross-diagnostic severity failures (e.g., malignant predicted as benign). This binary definition removes subjective ambiguity and provides a deterministic ground truth for training our error classifiers. Notice that humans may do some of the ISIC's {\em non-human errors}, but our methodology also works in this case, as we are targeting harmful errors.

\subsection{Base Classification}
\subsubsection{Model Architecture and Training}
The first stage involves a standard multi-class classification task. For the medical task, we employed a ResNet-50~\cite{he2016deep} architecture pre-trained on ImageNet ~\cite{deng2009imagenet,russakovsky2015imagenet} in PyTorch, following the benchmarks established
by~\cite{al2020multiple}. Although our technique is model-agnostic and any classifier can be used, to ensure consistent feature extraction capabilities, we employ the same ResNet-50 architecture as the backbone for the animal domain task as well.

Let $f_{base}(x)$ be the base classifier for an input image $x$, producing a predicted class $\hat{y}_{base}$ and a probability vector $\mathbf{p}$.
This model is trained using standard Cross-Entropy loss. We standardize on ResNet to facilitate direct comparison across varying complexity levels in the Oxford-IIIT Pet (37 classes), ISIC-2018 (7 classes), and SICAPv2 (4 classes) datasets. The Adam optimizer was used with a learning rate of 0.001.

\subsubsection{Animal Classification Task}
The first task used to test our methodology is fine-grained animal classification using 37 categories (12 cat breeds and 25 dog breeds) from the Oxford-IIIT Pet Dataset~\cite{parkhi2012cats}, which contains nearly 200 images per class, totaling 7,393 images, and serves as the training dataset. This setting presents challenges due to high inter-class visual similarity, varied error types (e.g., confusing species vs. confusing breeds), and implications for practical deployment scenarios where errors can carry different levels of severity.

For evaluation, we curated a separate test dataset of 18,500 images (500 per class) gathered from various web sources. This dataset shares the same class structure as the training set and is used to simulate a realistic deployment environment. Notice that this test dataset is much larger than the training set and probably more complex, making it harder to achieve an improvement.

\subsubsection{Skin Lesion Classification}
\sloppy
To validate our technique in a high-stakes medical domain, we use the ISIC-2018 dataset~\cite{codella2019skin,tschandl2018ham10000}. This dataset contains 10,015 dermoscopy images categorized into 7 diagnostic sub-classes: Melanoma (MEL), Melanocytic Nevus (NV), Basal Cell Carcinoma (BCC), Actinic Keratosis (AKIEC), Benign Keratosis (BKL), Dermatofibroma (DF), and Vascular Lesion (VASC). Crucially, these sub-classes map to two critical super-classes that determine patient outcome:
Benign (Safe): NV, BKL, DF, VASC. Malignant (Critical): MEL, BCC, AKIEC.
This hierarchy allows us to distinguish between minor diagnostic disagreements (confusing two benign or two malign types) and dangerous failures (confusing malignant with benign). For evaluation we used a 15\% random sample split (1,512 samples). The probability outputs generated by the base model on these test sets serve as the input for the subsequent error correction stages.

\subsubsection{Prostate Histopathology Classification}
To further validate the pipeline, we include the SICAPv2 dataset, a patch-based dataset for prostate cancer classification~\cite{silva2020going}. This dataset presents a four-class problem. We map Non-Cancerous (NC) tissue to the Benign superclass ($S=0$). Gleason Grades 3, 4, and 5, indicating an increase in cancer severity, are all assigned to the Malignant superclass ($S=1$). This mapping allows us to isolate critical non-human errors (confusing cancerous tissue with healthy tissue) from less severe human-like errors (confusing different grades of malignancy).

\subsection{Error Detection}
In a real-world setting, the system does not know if $\hat{y}_{base}$ is correct. Therefore, we introduce an intermediate binary classifier, the \textit{Error Detector}, to act as a {\em gatekeeper} and address {\bf RQ1}.
\begin{itemize}
    \item \textbf{Goal:} Predict whether the base multi-class model's prediction is Correct or Erroneous.
    \item \textbf{Input:} The feature vector for this stage is the probability distribution $\mathbf{p}$ produced by the base classifier. For training, we utilize the probability vectors generated by the base classifier on the training set. During inference, the classifier operates on the probability vectors of the test set.
    \item \textbf{Model and Training:} We utilize a Gradient Boosted Decision Tree (GBDT). The model is trained to learn the confidence patterns associated with successful and failed predictions directly from the training data distribution.
    \item \textbf{Labels:} Correct predictions from the base model are labeled $0$, and all misclassifications (regardless of severity) are labeled $1$.
    \item \textbf{Baseline Comparison:} To address error detection performance, we compare the GBDT against a Maximum Class Probability (MCP) baseline~\cite{hendrycks2016baseline}. The MCP flags a prediction as an error if the base model's top confidence score falls below the mean confidence threshold. In this gatekeeper stage, precision is the priority to ensure correct base predictions are not falsely flagged for correction.
\end{itemize}

\subsection{Error Type Classification}
Once a potential error is flagged by the gatekeeper, the system must determine the \textit{nature} of the error to decide on the mitigation strategy to address {\bf RQ2}.
\begin{itemize}
    \item \textbf{Goal:} Given that an error has been detected, predict if it is a {\em human-like} error (Label 0) or a {\em non-human} error (Label 1).
    \item \textbf{Input:} Similar to the previous step, this classifier uses the probability vectors generated by the base classifier. For training, we utilize the probability vectors of the specific samples in the training set where the base model made an error. During inference, the classifier operates on the probability vectors of the test set samples flagged as errors by the previous step.
    \item \textbf{Model and Training:} We utilize a second GBDT classifier. The model is trained on the isolated misclassified samples from the training output of the base model, enabling it to distinguish between the two error severity levels based on the specific signatures in the probability distribution.
    \item \textbf{Labels:} 
        \begin{itemize}
        \item \textit{Human-like (0):} $\hat{y}_{base}$ and $y_{true}$ belong to the same Superclass.
        \item \textit{Non-Human (1):} $\hat{y}_{base}$ and $y_{true}$ belong to different Superclasses.
    \end{itemize}
\end{itemize}

\subsection{Superclass Flip Policy}
When a {\em non-human} error is detected, we assume the model has correctly identified visual features but assigned them to the wrong semantic context (Superclass). The policy forces a switch to the alternative Superclass, mitigating the error harm.
For a binary Superclass problem (e.g., Cat vs. Dog, or Benign vs. Malignant):
\begin{equation}
    \hat{y}_{final} = \arg\max_{c \in S_{alt}} \mathbf{p}_c
\end{equation}
where $S_{alt}$ is the set of classes belonging to the Superclass opposite to $\hat{y}_{base}$, and $\mathbf{p}_c$ is the probability assigned by the base classifier. This effectively re-routes the prediction to the most confident class within the {\em correct} semantic category.

\subsection{Automated Mitigation Pipeline}
The final decision logic combines the output of the previous models to address {\bf RQ3}. Let $D(x)$ be the error detection prediction and $T(x)$ be the error type prediction. The final prediction $\hat{y}_{final}$ is determined as follows ($\hat{y}_{base}$ is the base classifier's predicted class):
\begin{enumerate}
    \item \textbf{Pass-through:} If $D(x) = 0$ (Predicted Correct), then $\hat{y}_{final} = \hat{y}_{base}$.
    \item \textbf{Safe Failure:} If $D(x) = 1$ (Predicted Error), we run the error type classifier obtaining $T(x)$. If $T(x) = 0$ (Predicted {\em human-like}), we retain the original prediction ($\hat{y}_{final} = \hat{y}_{base}$). This avoids over-correction on ambiguous samples where the base model's second guess might be worse.
    \item \textbf{Intervention:} If $D(x) = 1$ and $T(x) = 1$ (Predicted {\em Non-Human} Error), the \textit{Superclass Flip Policy} is triggered.
\end{enumerate}

To evaluate the practical deployment capability of this real-time pipeline, we measured the computational overhead added by the correction mechanism. The dual GBDT evaluation and policy application require a fixed 0.1 ms per image. Compared to the base CNN latencies, this results in a total inference time increase of 1.60\% for the animal dataset, 1.84\% for ISIC, and 1.70\% for SICAPv2, confirming minimal impact on system speed.

\subsection{Evaluation Metrics}
We evaluated performance through two complementary accuracy metrics designed to capture different aspects of classification quality. 
Class accuracy measures strict fine-grained matching (e.g., Persian to Persian or Melanoma to Melanoma), providing a precise assessment of recognition capabilities.

More critically, Super-class accuracy (Animal or Diagnostic accuracy) tracks high-level correctness (Cat vs. Dog or Benign vs. Malignant). This serves as our primary safety metric by directly quantifying reductions in high-consequence errors. These severe misclassifications (where models make fundamentally wrong categorical judgments) mirror dangerous failures in critical domains. By employing this dual-metric approach, we simultaneously assess both nuanced discrimination and the system's ability to avoid {\em non-human} errors in essential categorical distinctions.

Finally, to explicitly address class imbalance (a key concern in medical and fine-grained datasets), we rely on the Matthews Correlation Coefficient (MCC)~\cite{foody2023challenges}. Unlike accuracy, MCC generates a high score only if the classifier performs well across all confusion matrix categories, ensuring our reported gains are robust and not artifacts of the dominant class.

\section{Results}
\label{sec:results}
We evaluated the effectiveness of our fully automated correction pipeline on our three datasets. 
Unlike idealized "oracle" experiments, these results reflect the system's performance in a realistic setting where neither the presence nor the type of errors are known a priori.

\subsection{Animal Classification}

\subsubsection{Base Model Performance}
We first established the performance of the base ResNet-50 model. The model achieved a standard class accuracy of 76.86\% and an MCC of 0.76.
However, regarding semantic safety, the model achieved a Super-class Accuracy of 90.04\%, meaning that in nearly 10\% of cases (1,843 samples), the model confused a cat with a dog or vice versa, a {\em non-human} error.

\begin{figure}[t]
    \centering
    \vspace*{-0.3cm}
    \begin{minipage}{0.58\textwidth}
        \centering
        \includegraphics[width=\linewidth]{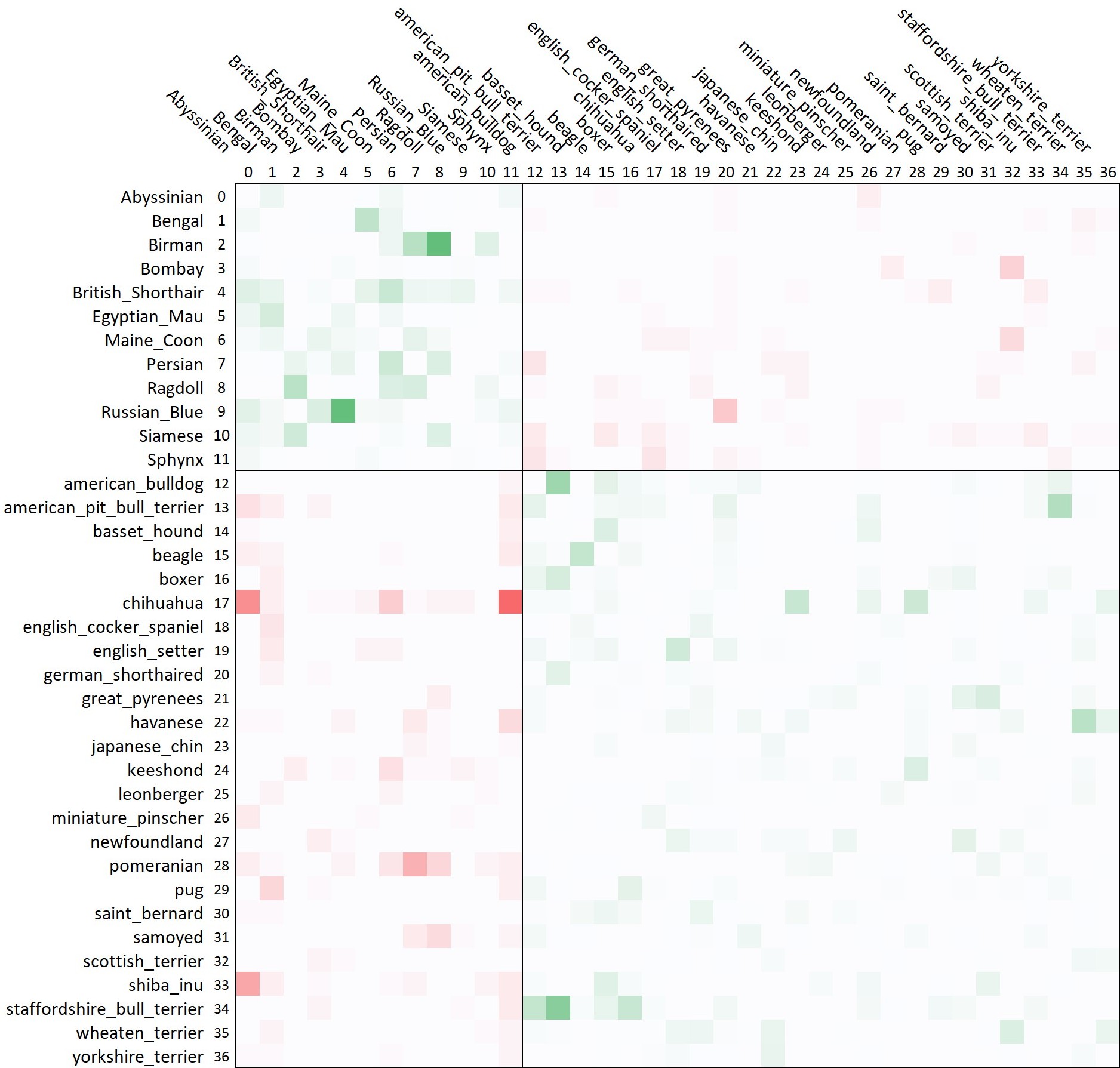}
        \caption{Confusion Matrix (Animals Dataset), distinguishing cat (0--11) and dog (12--36) breeds.}
        \label{fig:confusion}
    \end{minipage}
    \hfill
    \begin{minipage}{0.38\textwidth}
        \centering
        \fbox{\includegraphics[width=0.85\linewidth]{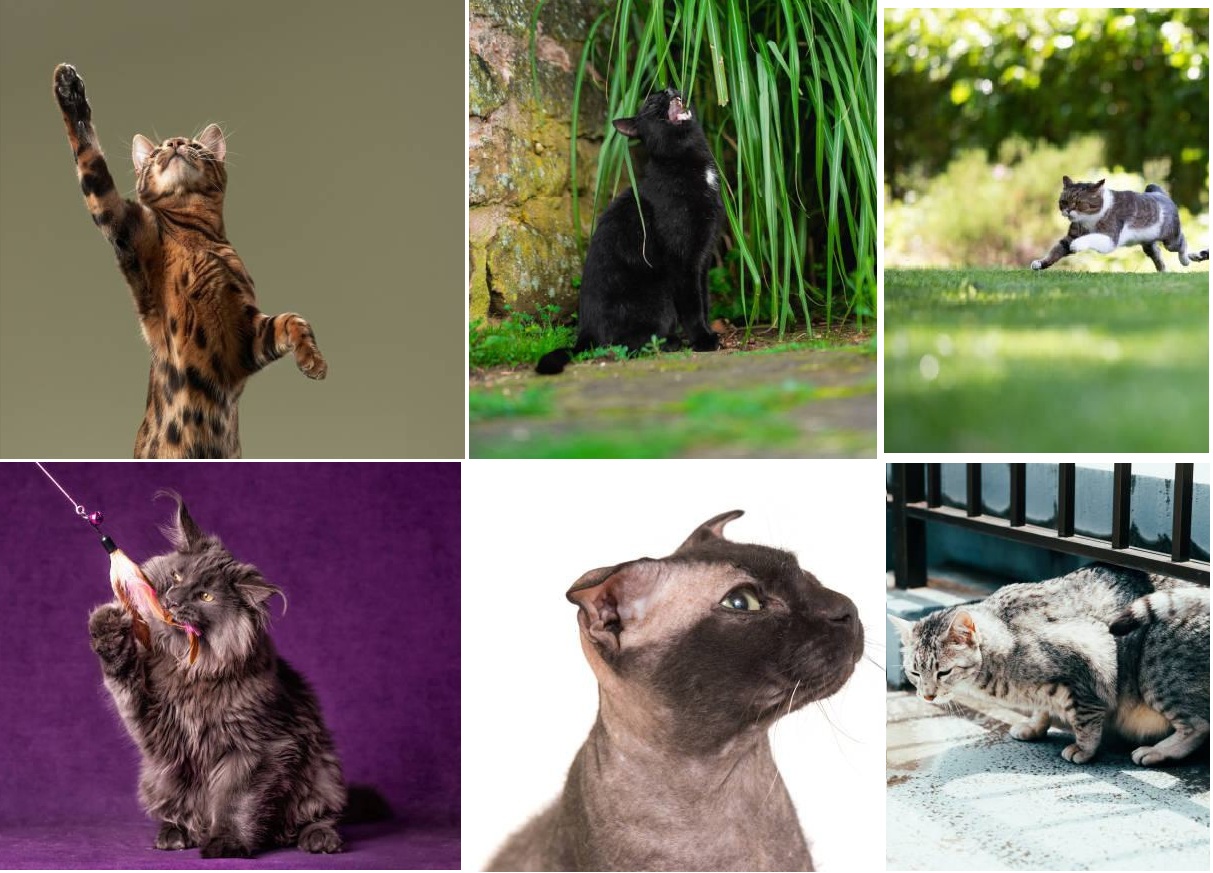}}
        \\[0.1cm] 
        \fbox{\includegraphics[width=0.85\linewidth]{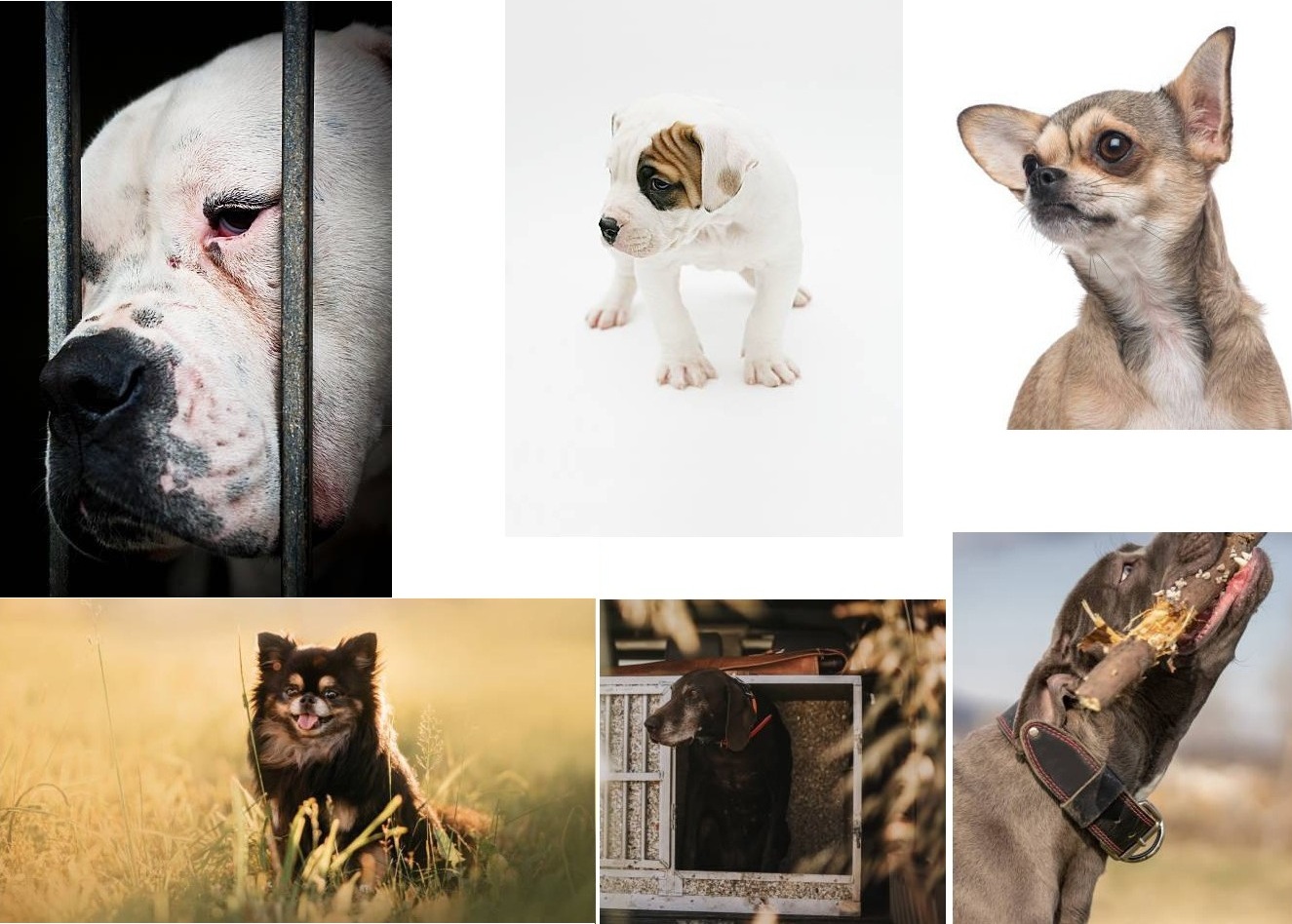}}
        \caption{Non-human errors: Cats misclassified as dogs (top) and dogs as cats (bottom).}
        \label{fig:non_human}
    \end{minipage}
    \vspace*{-0.4cm}
\end{figure}

Error samples ($n=4,280$) are detailed in Figure~\ref{fig:confusion}, which displays the confusion matrix of the base classifier on the test dataset. In this matrix, rows represent the true labels (True Cat, True Dog) and columns represent the model's predictions (Predicted Cat, Predicted Dog). Darker squares indicate a higher number of misclassifications.

Out of the total errors, 1,843 were identified as {\em non-human} (severe) errors. Specifically, the model incorrectly predicted 848 cats as dogs (top right) and 995 dogs as cats (bottom left). 
The remaining 2,437 errors were categorized as {\em human-like} errors (green areas). These consist of 650 instances where cats were misclassified as other cat breeds (top left) and 1,787 instances where dogs were misclassified as other dog breeds (bottom right). 
Qualitative examples of {\em non-human} errors are shown in Figure~\ref{fig:non_human} \
. These results highlight that while {\em human-like} errors are more frequent, {\em non-human} errors constitute a measurable portion of failures that require targeted correction.

\subsubsection{Mitigation Pipeline Performance}

The Error Detector model achieved a MCC of 0.32, while the Error Type Classifier achieved a MCC of 0.26. When compared to the MCP baseline, which yielded an F1 score of 0.61 and a precision of 0.51, our GBDT error detector prioritized a conservative approach, achieving a precision of 0.62 with an F1 score of 0.39. This higher precision ensures that correct base predictions are preserved. 

While these scores indicate the difficulty of the task, they provided sufficient signal for the policy to operate effectively, reducing the count of {\em non-human} errors from 1,843 to 1,702 (an improvement of 7.7\%).

The overall class accuracy increased to 77.64\% while the animal accuracy increased from 90.04\% to 90.80\% (almost 0.8\% increase in both cases). The number of correct predictions increased from 14,220 to 14,363, representing a net recovery of 143 images. This confirms that by targeting {\em non-human} errors, we not only improved safety but also correctly reclassified samples that were previously wrong. All these results are given in Table~\ref{tab:animal_results}, while Table~\ref{tab:animal_cm} shows the shift in the error distribution.
\begin{table}[t]
    \centering
    \begin{minipage}{0.48\textwidth}
        \centering
        \caption{Performance Comparison on Animal Dataset (Base Model vs. Pipeline).}
        \label{tab:animal_results}
        \resizebox{\textwidth}{!}{%
        \begin{tabular}{|l|c|c|c|}
        \hline
        \textbf{Method} & \textbf{Base} & \textbf{Pipeline} & \textbf{$\Delta$ (\%)} \\
        \cline{1-1}
        \textbf{Metric} & \textbf{Model} &  &  \\
        \hline
        \# Correct & 14,220 & 14,363 & 1.00\% \\
        \textbf{NH errors}~~ & \textbf{1,843} & \textbf{1,702} & \textbf{-7.65\%} \\
        HL errors & 2,437 & 2,435 & -0.1\% \\ \hline
        Class Acc. & 76.86\% & 77.64\% & 1.01\% \\
        \textbf{Animal Acc.} & \textbf{90.04\%} & \textbf{90.80\%} & \textbf{0.84\%} \\
        MCC & 0.763 & 0.771 & 1.0\% \\
        \hline
        \end{tabular}}
    \end{minipage}\hfill
        \begin{minipage}{0.48\textwidth}
        \centering
        \caption{Performance Comparison on ISIC-2018 (Base Model vs. Pipeline).}
        \label{tab:isic_results}
        \resizebox{\textwidth}{!}{%
        \begin{tabular}{|l|c|c|c|}
        \hline
        \textbf{Method} & \textbf{Base} & \textbf{Pipeline} & \textbf{$\Delta$ (\%)} \\ \cline{1-1}
        \textbf{Metric} & \textbf{Model} &  &  \\
        \hline
        \# Correct & 1,198 & 1,262 & 5.3\% \\
        \textbf{NH Errors} & \textbf{220} & \textbf{145} & \textbf{-34.1\%} \\
        HL Errors & 94 & 105 & 11.7\% \\ \hline
        Class Accuracy & 79.23\% & 83.47\% & 5.35\% \\
        \textbf{S. Class Acc.} & \textbf{85.45\%} & \textbf{90.41\%} & \textbf{5.80\%} \\
        MCC & 0.667 & 0.723 & 8.4\% \\
        \hline
        \end{tabular}}
    \end{minipage}\hfill
    \vspace*{-0.2cm}
\end{table}

\begin{table}[t]
    \centering
 \begin{minipage}{0.48\textwidth}
        \centering
        \caption{Breakdown of errors (Animal Dataset). HL: {\em Human-like}, NH: {\em Non-Human}.}
        \label{tab:animal_cm}
        \resizebox{\textwidth}{!}{%
        \begin{tabular}{|l|c|c|c|c|}
        \hline
        & \multicolumn{2}{c|}{\textbf{Base Model}} & \multicolumn{2}{c|}{\textbf{Pipeline}} \\
        \hline
        \textbf{Metric} & \textbf{Cat} & \textbf{Dog} & \textbf{Cat} & \textbf{Dog} \\
        \hline
        \textbf{True Cat} & 650 (HL) & 848 (NH) & 676 (HL) & 705 (NH) \\
        \hline
        \textbf{True Dog} & 995 (NH) & 1,787 (HL) & 997 (NH) & 1,759 (HL) \\
        \hline
        \end{tabular}}
    \end{minipage}
    \begin{minipage}{0.48\textwidth}
        \centering
        \caption{Breakdown of Errors for ISIC 2018. HL: {\em Human-like}, NH: {\em Non-human}.}
        \label{tab:isic_cm}
        \resizebox{\textwidth}{!}{%
        \begin{tabular}{|l|c|c|c|c|}
        \hline
        & \multicolumn{2}{c|}{\textbf{Base Model}} & \multicolumn{2}{c|}{\textbf{Pipeline}} \\
        \hline
        \textbf{Metric} & \textbf{Ben.} & \textbf{Mal.} & \textbf{Ben.} & \textbf{Mal.} \\
        \hline
        \textbf{True Ben.} & 64 (HL) & 153 (NH) & 78 (HL) & 75 (NH) \\
        \hline
        \textbf{True Mal.} & 67 (NH) & 30 (HL) & 70 (NH) & 27 (HL) \\
        \hline
        \end{tabular}}
    \end{minipage}
    \vspace*{-0.2cm}
\end{table}

\subsection{Skin Lesion Classification}
\subsubsection{Base Model Performance}
The base model on the ISIC dataset achieved a fine-grained accuracy of 79.23\% and a Super-class (Diagnostic) Accuracy of 85.45\%.
Analysis of the 314 error samples revealed a dangerous trend: 220 errors (70.0\%) were {\em non-human} (Critical), involving the confusion of Benign and Malignant lesions. Specifically, 153 Benign cases were misclassified as Malignant, and 67 Malignant cases were missed as Benign. The remaining 94 errors (30.0\%) were {\em human-like}, involving confusion within the same diagnostic category (e.g., confusing two types of benign lesions).

\subsubsection{Mitigation Pipeline Performance}
The Error Detector achieved an accuracy of 83.20\% and a MCC of 0.42. Compared to the MCP baseline (precision 0.446, F1 0.550), the GBDT detector maintained higher precision (0.67) with an F1 score of 0.49. The Error Type Classifier achieved an accuracy of 88.43\% with a strong MCC of 0.70, indicating a high capability to distinguish between mild and critical errors. Indeed, the number of {\em non-human} errors was reduced by 34.1\%, dropping from 220 to 145, leading to a 5.80\% improvement in Super-class Accuracy (Diagnostic Safety) from 85.45\% to 90.41\%. \

As detailed in Table~\ref{tab:isic_results}, overall fine-grained accuracy improved from 79.23\% to 83.47\%. Critically, this improvement corresponds to an increase in Correct Predictions from 1,198 to 1,262. This represents 64 additional patients correctly diagnosed, a substantial gain given the small size of the test set. Table~\ref{tab:isic_cm} shows the detailed confusion matrix.



\subsection{Prostate Histopathology Classification}
\subsubsection{Base Model Performance}
The base model on the SICAPv2 dataset achieved an accuracy of 68.43\% and an MCC of 0.5489. Regarding semantic safety, the model recorded a Super-class Accuracy of 91.00\%. The base model generated 191 {\em non-human} errors, representing cases where malignant tissue was confused with benign tissue, or vice versa. The model also produced 479 {\em human-like} errors, which involve misclassifications among the different Gleason grades of malignancy.

\subsubsection{Mitigation Pipeline Performance}
The Error Detector for SICAPv2 achieved an accuracy of 68.28\% and an MCC of 0.09. In comparison to the MCP baseline (precision 0.48, F1 0.51), the GBDT detector maintained a similar precision of 0.49 but with a lower F1 score of 0.16, indicating a highly conservative correction gating. The Error Type Classifier, however, demonstrated robust performance with an accuracy of 91.94\% and an MCC of 0.84.

Applying the final correction policy successfully reduced the {\em non-human} error count from 191 to 167 (a 12.57\% reduction). This targeted mitigation improved the Super-class Accuracy to 92.13\% and the overall fine-grained accuracy to 69.18\%. 
Table~\ref{tab:sicap_results} summarizes the performance gains, and Table~\ref{tab:sicap_cm} provides the corresponding error breakdown.

\begin{table}[t]
    \centering
    \begin{minipage}{0.48\textwidth}
        \centering
        \caption{Performance Comparison on SICAPv2 (Base Model vs. Pipeline).}
        \label{tab:sicap_results}
        \resizebox{\textwidth}{!}{%
        \begin{tabular}{|l|c|c|c|}
        \hline
        \textbf{Method} & \textbf{Base} & \textbf{Pipeline} & \textbf{$\Delta$ (\%)} \\ \cline{1-1}
        \textbf{Metric} & \textbf{Model} &  & \\
        \hline
        \# Correct & 1,452 & 1,468 & 1.10\% \\
        \textbf{NH Errors} & \textbf{191} & \textbf{167} & \textbf{-12.57\%} \\
        HL Errors & 479 & 487 & 1.67\% \\ \hline
        Class Accuracy & 68.43\% & 69.18\% & 1.10\% \\
        \textbf{S. Class Acc.} & \textbf{91.00\%} & \textbf{92.13\%} & \textbf{1.24\%} \\
        MCC & 0.5489 & 0.5596 & 1.95\% \\
        \hline
        \end{tabular}}
    \end{minipage}\hfill
    \begin{minipage}{0.48\textwidth}
        \centering
        \caption{Breakdown of Errors for SICAPv2. HL: {\em Human-like}, NH: {\em Non-human}.}
        \label{tab:sicap_cm}
        \resizebox{\textwidth}{!}{%
        \begin{tabular}{|l|c|c|c|c|}
        \hline
        & \multicolumn{2}{c|}{\textbf{Base Model}} & \multicolumn{2}{c|}{\textbf{Pipeline}} \\
        \hline
        \textbf{Metric} & \textbf{Ben.} & \textbf{Mal.} & \textbf{Ben.} & \textbf{Mal.} \\
        \hline
        \textbf{True Ben.} & 0 (HL) & 55 (NH) & 0 (HL) & 52 (NH) \\
        \hline
        \textbf{True Mal.} & 136 (NH) & 479 (HL) & 115 (NH) & 487 (HL) \\
        \hline
        \end{tabular}}
    \end{minipage}
    \vspace*{-0.2cm}
\end{table}

\section{Discussion}
\label{sec:discussion}

\subsection{Real-World Applicability and Robustness}
A major challenge in deploying error correction systems is the absence of a perfect oracle to signal when a model has failed. Our pipeline addresses this by learning to detect errors directly from the model's latent representations. As observed in our results, this detection step is not infallible; the error detection classifier achieved a Matthews Correlation Coefficient (MCC) ranging between 0.09 and 0.42 across the three evaluated datasets.

Unlike standard confidence calibration, trust score methods~\cite{jiang2018trust}, or the evaluated Maximum Class Probability (MCP) baseline, which only flag uncertain samples, our pipeline goes a step further by actively mitigating them. While trust scores and MCP can identify potential failures, they offer no mechanism to repair the prediction.

This performance level implies a necessary trade-off: the introduction of noise in the form of False Positives, where the system flags correct predictions as potential errors. As shown by our baseline comparisons, prioritizing recall in this stage introduces excessive false corrections. By prioritizing precision, our framework mitigates this risk. A key finding of this study is the method's robustness to this noise. Despite the imperfect detection of errors, the mitigation pipeline results show an improvement in both accuracy and semantic safety compared to the baseline. This proves that the correction policy is sufficiently conservative and precise; even when the system must estimate which samples are wrong, it repairs more errors than it introduces, resulting in a net gain in reliability.

\subsection{Error Reduction Dynamics}
Analyzing the shifts in error categories reveals important nuances in how safety is improved. For the ISIC and SICAPv2 datasets, while the total number of errors dropped, the specific count of {\em human-like} errors saw an increase(from 94 to 105 in ISIC, and 479 to 487 in SICAPv2) in the final output. This increase represents a positive outcome indicating that the policy successfully intercepted severe {\em non-human} errors, such as a malignant lesion being misclassified as benign, and, in cases where it could not fully correct them, shifted them into the {\em human-like} category (e.g., predicting a different malignant subclass or a different Gleason grade). By effectively converting dangerous failures into mild and intelligible mistakes, the system minimizes the potential harm of misclassifications, which is a critical requirement for medical diagnostic support.

\subsection{Limitations}
While the proposed method has demonstrated improvements, several challenges remain. First, the performance of the error classifier indicates room for improvement; enhancing classification accuracy may require deeper feature representation or integrating meta-learning techniques to strengthen generalization. 
Second, the task-specific nature of error definitions requires domain expertise for categorizing {\em non-human} errors, limiting adaptability across different contexts without manual intervention. 
Third, scalability concerns persist, as balancing rare error types in larger datasets remains challenging. 
Fourth, we acknowledge that representation learning methods like CLIP~\cite{radford2021learning} and DINO~\cite{caron2021emerging} may provide fundamental solutions through better embeddings. However, our post-hoc approach offers practical value in resource-constrained scenarios.
Finally, a fundamental limitation is the reliance on outcome-based error classification (e.g., breed vs. species confusion). This approach does not fully capture the behavioral differences in decision-making between humans and models, such as the reliance on distinct image textures versus shapes.
\section{Conclusions and Future Work}
\label{sec:Conclusion}

In this work we introduced a novel technique to align deep learning classification errors with human cognitive patterns. By distinguishing between {\em human-like} (semantically plausible) and {\em non-human} (severe/illogical) errors, we designed a pipeline that detects and corrects these failures without altering the backbone model. Our empirical results validate the effectiveness of this approach in three use cases and positively answer our three research questions. Furthermore, our evaluations confirm the system's viability for practical deployment, demonstrating a latency overhead of under 2\% across all tasks and achieving higher error detection precision than standard Maximum Class Probability baselines.

In the real-world animal case, the pipeline reduced severe {\em non-human} errors by 7.7\%, while increasing the final accuracy to 77.64\%. Similarly, in the medical case of ISIC 2018, {\em non-human} errors were reduced by 34.1\%, boosting the super-class safety accuracy to 90.41\%. On the SICAPv2 prostate histopathology dataset, the framework decreased critical {\em non-human} errors by 12.57\%, raising the super-class safety accuracy to 92.13\%. Crucially, the results on the medical datasets showed that even when errors persisted, the system often shifted them from dangerous {\em non-human} misclassifications to safer {\em human-like} ones, thereby enhancing the trustworthiness of the system in high-stakes environments. 

Notice that our technique depends on having a multi-class setting, so we can automatically detect non-human errors. Our pipeline even beats a binary super-class classifier. For example, for ISIC 2018, using the same base model (ResNet-50), we obtain an accuracy of 89.4\% (so a 1.2\% improvement) with lower MCC (0.67). 

To address the current limitations, future research will explore hybrid correction policies that integrate uncertainty-based human delegation for ambiguous cases. We also aim to evaluate cross-domain generalization in fields like Natural Language Processing, where factual hallucinations present similar challenges. To reduce manual annotation, future frameworks could dynamically learn error taxonomies through unsupervised clustering or behavior-based analysis (e.g., using attention maps). Finally, incorporating cost-sensitive learning, adaptive thresholding, and confidence calibration~\cite{guo2017calibration} will further optimize the correction gating, ensuring robust performance in imbalanced and safety-critical environments.

\bibliographystyle{splncs04}
\bibliography{bib}

\end{document}